\documentclass[11pt]{article}

\usepackage[preprint]{acl}
\usepackage{times}
\usepackage{latexsym}
\usepackage{amsmath}
\usepackage[T1]{fontenc}

\usepackage[utf8]{inputenc}

\usepackage{microtype}

\usepackage{inconsolata}

\usepackage{times}
\usepackage{latexsym}

\usepackage{graphics}
\usepackage{graphicx}
\usepackage{color}
\usepackage{colortbl}
\usepackage{amsfonts}
\usepackage{amssymb}
\usepackage{amsmath}
\usepackage{nicefrac}
\usepackage{bbm}
\usepackage{booktabs}
\usepackage{microtype}
\usepackage{multirow}
\usepackage{multicol}
\usepackage{tabu}
\usepackage{amsthm}
\usepackage{footmisc}
\usepackage{soul}
\usepackage{arydshln}
\usepackage{wasysym}
\usepackage{array}
\usepackage{makecell}
\usepackage{scalerel}

\usepackage{algpseudocode}
\usepackage{algorithm}
\usepackage{hyperref}
\usepackage{xspace}
\usepackage{url}
\usepackage{utfsym}
\usepackage{dirtytalk}

\usepackage{float}
\usepackage{enumitem}
\usepackage{tikz}
\usepackage{tcolorbox}
\usepackage{siunitx}

\usepackage{subcaption} 

\newcolumntype{L}[1]{>{\let\newline\\\arraybackslash\hspace{0pt}}m{#1}}

\title{Culturally Grounded Physical Commonsense Reasoning in Italian and English: A Submission to the MRL 2025 Shared Task
}

\author{Marco De Santis \\
  University of Udine \\
  \texttt{desantis.marco001@spes.uniud.it} \\\And
  Lisa Alazraki \\
  Imperial College London \\
  \texttt{lisa.alazraki20@imperial.ac.uk} \\}

\begin{document}
\maketitle
\begin{abstract}
This paper presents our submission to the MRL 2025 Shared Task on Multilingual Physical Reasoning Datasets. The objective of the shared task is to create manually-annotated evaluation data in the physical commonsense reasoning domain, for languages other than English, following a format similar to PIQA \cite{Bisk_Zellers_Lebras_Gao_Choi_2020}. Our contribution, \mbox{\textsc{FormaMentis}}\footnote{Data available upon request.}, is a novel benchmark for physical commonsense reasoning that is grounded in Italian language and culture. The data samples in \textsc{FormaMentis} are created by expert annotators who are native Italian speakers and are familiar with local customs and norms. The samples are additionally translated into English, while preserving the cultural elements unique to the Italian context. 
\end{abstract}

\section{Introduction}

Commonsense reasoning in language models is an active area of research \cite{kavumba-etal-2019-choosing, 2021_3ef81541, gupta-etal-2023-john, li2025hellaswagprolargescalebilingualbenchmark}, with recent work extending beyond English to other languages \cite{ghosh2025multilingualmindsurvey}. Existing multilingual datasets, however, have primarily targeted causal reasoning \cite{ponti-etal-2020-xcopa}, sentence completion, and general question answering \cite{lin-etal-2021-common}, leaving other aspects of commonsense underexplored. In this work, we highlight physical commonsense as a particularly important yet overlooked dimension for multilingual research. Unlike causal or textual reasoning, physical commonsense requires reasoning over objects and actions that are often grounded in culture-specific practices, and may even lack direct English equivalents \cite{anacleto2006using, zou2009culture}. 

\begin{figure}[!t]
    \centering
    \includegraphics[width=\linewidth]{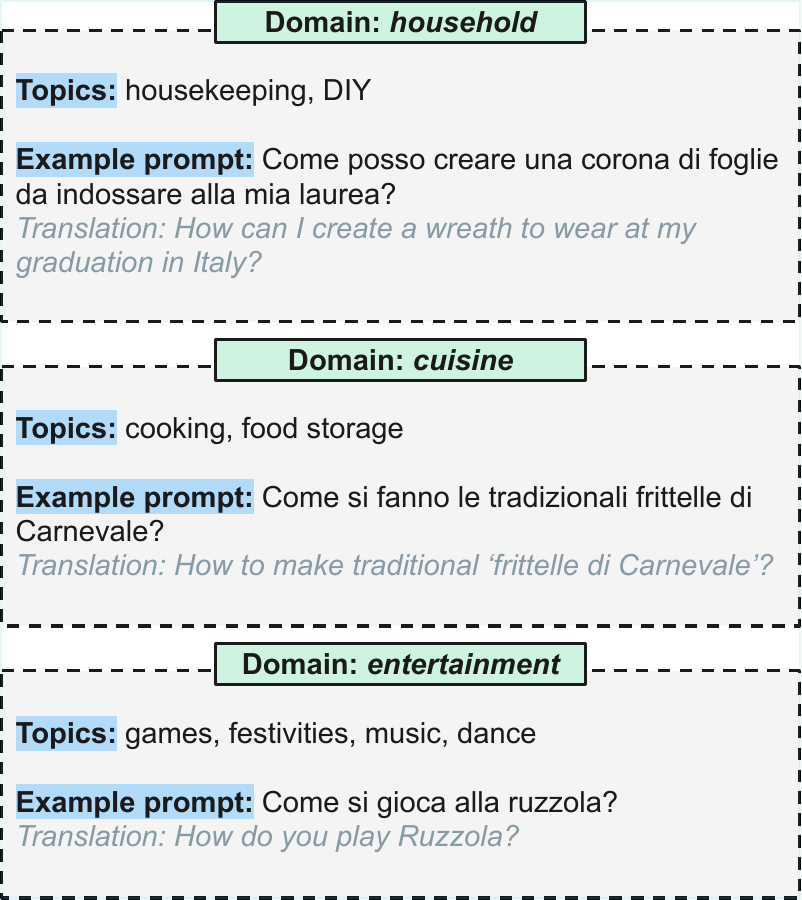} \vspace{-16pt}
    \caption{\textsc{FormaMentis} domains. For each, we show an example prompt requiring cultural knowledge. E.g., a graduation wreath is a specific type of wreath worn by Italian university students, \textit{frittelle di Carnevale} is a local holiday recipe, and \textit{Ruzzola} is a traditional Italian country sport. Completing these prompts correctly thus requires an understanding of the specific characteristics and rules of each item or practice. It is also worth noting that the English translations aim to preserve these culture-specific characteristics, which involves leaving words in their original Italian form where necessary.}
    \label{fig:main}
\end{figure}

Building on the above observations, we introduce \textsc{FormaMentis}, a new Italian-language benchmark manually constructed by expert annotators who are native speakers. \textsc{FormaMentis} focuses on everyday scenarios involving objects and actions rooted in Italian customs. The samples in the dataset are distributed among three domains (\textit{household}, \textit{cuisine}, and \textit{entertainment}) that require reasoning about physical items and practices presupposing cultural knowledge. To broaden accessibility and facilitate cross-lingual evaluation, all samples are also manually translated into English, with careful attention to preserving the Italian-specific cultural references embedded in the original.

\section{The \textsc{FormaMentis} Benchmark}

\subsection{Data Format}

The \textsc{FormaMentis} benchmark adopts a format similar to the PIQA dataset \cite{Bisk_Zellers_Lebras_Gao_Choi_2020}: each sample consists of a prompt paired with two candidate completions, only one of which is correct. The completions are closely matched, typically differing by just one or two words, with the incorrect choice designed to be clearly wrong but not so implausible as to make the task trivial.

\subsection{Data Collection}

The data samples in \textsc{FormaMentis} are manually created by expert annotators who are native Italian speakers, with detailed guidelines serving as a reference (see Appendix~\ref{sec:guidelines}). Every sample must be novel (i.e., not translated from other sources) and must belong to one of three domains reflecting local customs: \textit{household}, \textit{cuisine}, and \textit{entertainment}. Figure~\ref{fig:main} presents an overview of these domains, their associated topics, and representative prompts.

Annotators are instructed to create samples that demand physical commonsense reasoning and incorporate cultural references rooted in Italian everyday culture. These include activities tied to local traditions as well as linguistic expressions that do not translate directly into other languages.

\subsection{Data Validation}

Each data sample in \textsc{FormaMentis} undergoes evaluation through a multi-step validation questionnaire, carried out by a native speaker other than the sample author. The validation steps are detailed in Table~\ref{tab:val}. For step \#5, minor adjustments to determiners, verbs, adjectives, or pronouns are not counted as substantive changes when they arise solely as a consequence of substituting other words. This exception reflects the rules of Italian grammar, which require agreement in gender (feminine/masculine) and number (singular/plural).

If any validation step receives a negative outcome, the sample must be revised or rewritten, after which the full assessment is repeated. Only those samples that successfully pass every step of the questionnaire are included in the final benchmark.

\begin{figure}[!t]
    \centering
    \includegraphics[width=\linewidth]{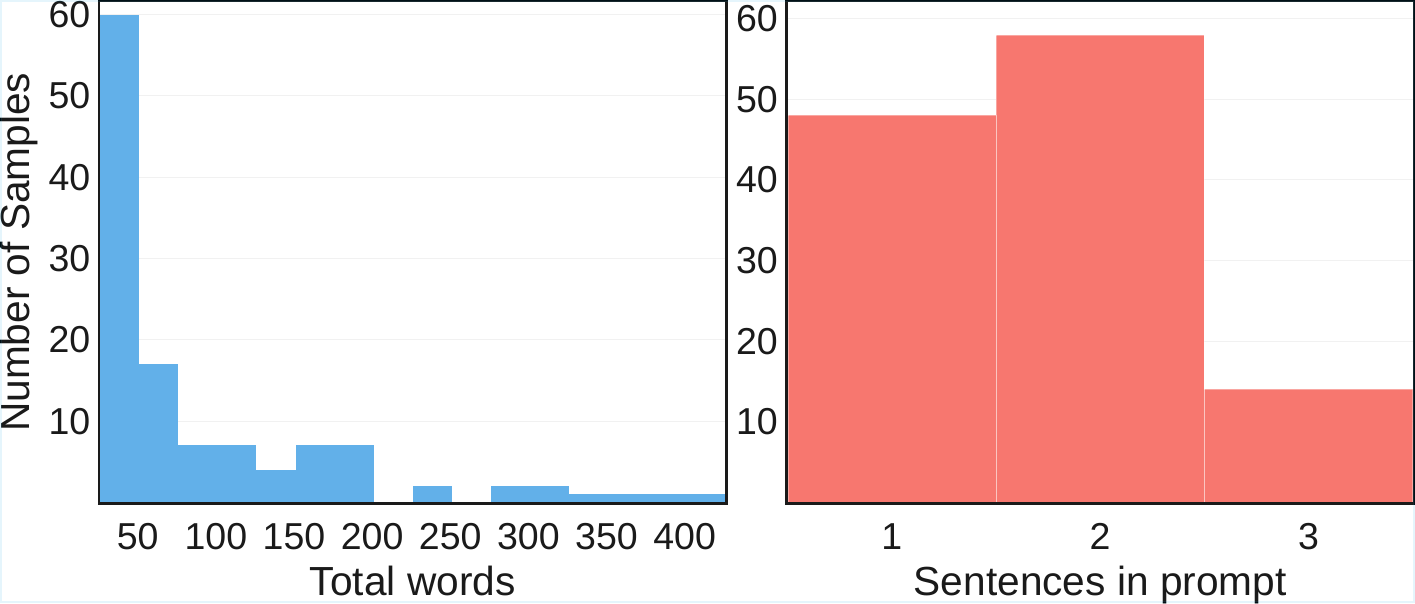}\vspace{-6pt}
    \caption{Sample distribution by total number of words (left) and number of sentences in the prompt (right).}
    \label{fig:hist}
\end{figure}

\begin{table}[!t]
\centering
\renewcommand\arraystretch{1.0}
\setlength{\tabcolsep}{1pt}
\scalebox{0.77}{
\begin{tabular}{L{1.3cm}L{8.2cm}}
\hline
\toprule
\textbf{Step} & \textbf{Validation Question} \\
\midrule
{\#1} & {Does the prompt require physical reasoning and knowledge beyond what is stated in it? [Yes/No]} \\
\midrule
{\#2} & {Is the physical knowledge required common among native Italian speakers? [Yes/No]} \\
\midrule
{\#3} & {Do the prompt and/or completions contain cultural references, linguistic expressions or colloquialisms that are specific to Italian culture? [Yes/No]} \\
\midrule
{\#4} & {Is the prompt unambiguous, and is only one of the two candidate completions correct? [Yes/No]} \\
\midrule
{\#5} & {Are the two candidate completions mostly similar, differing only by one or two words? [Yes/No] } \\
\midrule
{\#6} & {Is the incorrect completion plausible enough not to appear absurd? [Yes/No]} \\
\bottomrule
\hline
\end{tabular}
}\vspace{-3pt}
\caption{
Validation steps for all samples in \textsc{FormaMentis}. Only samples that obtain a positive answer at all steps are included in the benchmark.
}
\label{tab:val}

\end{table}

\subsection{Data Statistics}\label{sec:stats}

\paragraph{Number of samples.} \textsc{FormaMentis} contains 120 test samples, a size consistent with other high-quality, human-written reasoning benchmarks \cite{app13042577, press-etal-2023-measuring, 10.1162/dint_a_00234, alazraki2025agentcomacompositionalbenchmarkmixing}. The samples are evenly distributed across the three domains -- household, cuisine, and entertainment -- with 40 samples per domain.\vspace{-5pt}
\paragraph{Sample length.} The total sample length (Italian-language prompt + solution) ranges from 26 to 452 words, with half of the samples falling in the 26–50 bin, as shown in Figure~\ref{fig:hist}. Sixty percent of the samples feature multi-sentence prompts (58 contain two sentences and 14 contain three), while the rest (40\%) are single-sentence prompts. Additional data statistics are provided in Appendix~\ref{sec:further_stats}.

\section{Conclusion}

We introduced \textsc{FormaMentis}, a manually curated benchmark for physical commonsense reasoning in Italian, focused on everyday scenarios reflecting local practices. Our benchmark is carefully validated for quality and includes English translations that preserve contextual nuances. By providing this resource, we aim to support research on multilingual physical reasoning and facilitate the evaluation of language models on culturally grounded commonsense tasks beyond English.

\section*{Limitations}

\textsc{FormaMentis} is created for the text-based evaluation of culturally grounded physical commonsense reasoning in language models. As such, the benchmark does not contain any images or videos, though we acknowledge that such multimodal inputs can support physical reasoning. We leave the study of multimodal physical reasoning that integrates culture-specific knowledge to future work.

\section*{Ethical Considerations}

All samples in \textsc{FormaMentis} are manually written by human experts, which guarantees originality and independence from third-party licensed sources. We have carefully reviewed the dataset to avoid any offensive or harmful content, and all samples focus on neutral, everyday scenarios.

\bibliography{custom} \clearpage

\appendix

\section{Annotator Guidelines}
\label{sec:guidelines}

The expert annotators follow the guidelines below when creating the samples in \textsc{FormaMentis}. All samples are written in Italian and are subsequently manually translated into English, with care taken to preserve the cultural cues and linguistic expressions of the original.

\bigskip

\noindent\small{\texttt{\textbf{FormaMentis: A Culturally Grounded Physical Commonsense Reasoning Benchmark in Italian.}}}

\bigskip

\noindent{\small{\texttt{Instructions (see also the original Shared Task instruction slides).\\ 
\\
Note: Samples must be manually created. The use of LLMs to write the samples is *not* allowed.\\
You may refer to the PIQA dataset for the format of questions/answers required, but please *do not* translate the samples from PIQA (or elsewhere).
\\
\\
The prompt+completions samples need to be about:
\\
- Physical reasoning: For each example, the solution must relate to physical properties of one or more objects. You may include common physical tasks or actions.
\\
- Common sense: For each example, an average person who speaks your language natively should know the answer.
\\
\\
They also need to be culturally specific:
\\
For example, some items may not be easily translatable into English, or may require regional and/or cultural commonsense.
\\
\\
Other requirements: 
\\
\\
Use items of variable length. Try not to include too many short items, as they may be too easy for larger models. If possible:\\
\\
- Examples (prompt+solution) should be over 25 words long.\\
\\
- There should be some prompts that are multiple sentences long.\\
\\
- The two candidate solutions should be as similar as possible (e.g. differing only by one or two words, or just flipping the order of two phrases). One solution should be unambiguously correct and the other incorrect.\\
\\
- To ensure that the benchmark is not too "easy", the incorrect solution should not be so absurd that it is extremely obvious.\\
\\
- Try not to start all examples the same way.
\\
\\
Please create an equal number of samples in each of the three categories:
\\
\\
- Household (includes housekeeping and DIY)\\
- Cuisine (includes cooking and food storage)\\
- Entertainment (includes games, festivities, music and dance)}

\normalsize

\section{Fine-grained Data Statistics}\label{sec:further_stats}

\subsection{Length of Prompts}

The distribution of prompt lengths in \textsc{FormaMentis} is shown in Figure~\ref{fig:prompts}. Lengths range from 4 to 61 words, with most prompts containing fewer than 25. Word counts are obtained by splitting text sequences at whitespace. In Italian, words may be abbreviated and joined with an apostrophe. Although such pairs are grammatically distinct words, we treat them as a single unit in our analysis.

\subsection{Length of Completions}

Figure~\ref{fig:completions} presents the distribution of completion lengths in \textsc{FormaMentis}. These cover a wide range from 2 to 223 words, with approximately half (62 completions) containing fewer than 15 words. Because the two completion options are always identical or nearly identical in length, we measure the aggregate length of both options for each sample and divide this value by two to obtain the length of an individual option.

\begin{figure}[!b]
    \centering
    \includegraphics[width=\linewidth]{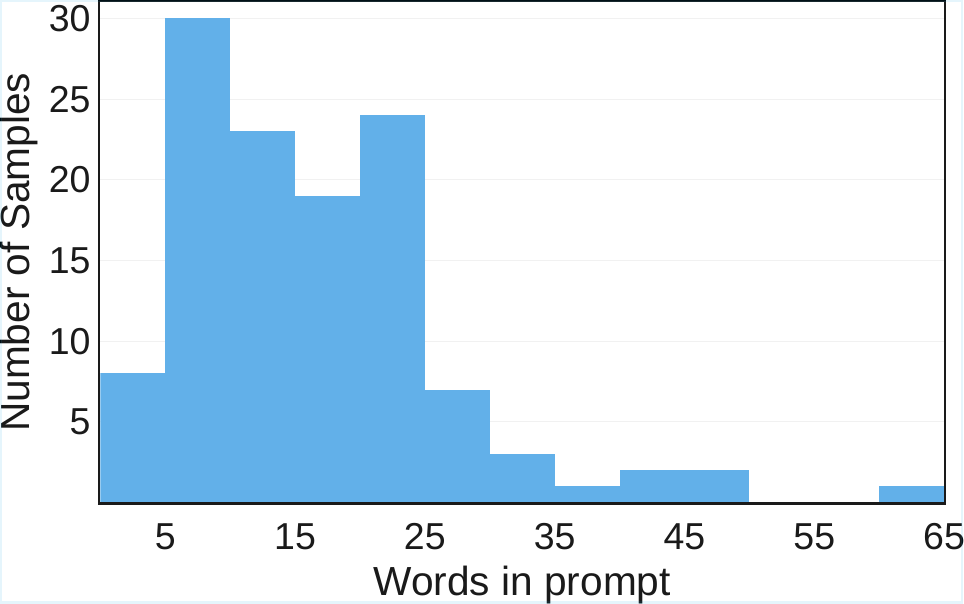}
    \caption{Sample distribution in \textsc{FormaMentis} by number of words in the prompt.}
    \label{fig:prompts}
\end{figure}

\begin{figure}[!b]
    \centering
    \includegraphics[width=\linewidth]{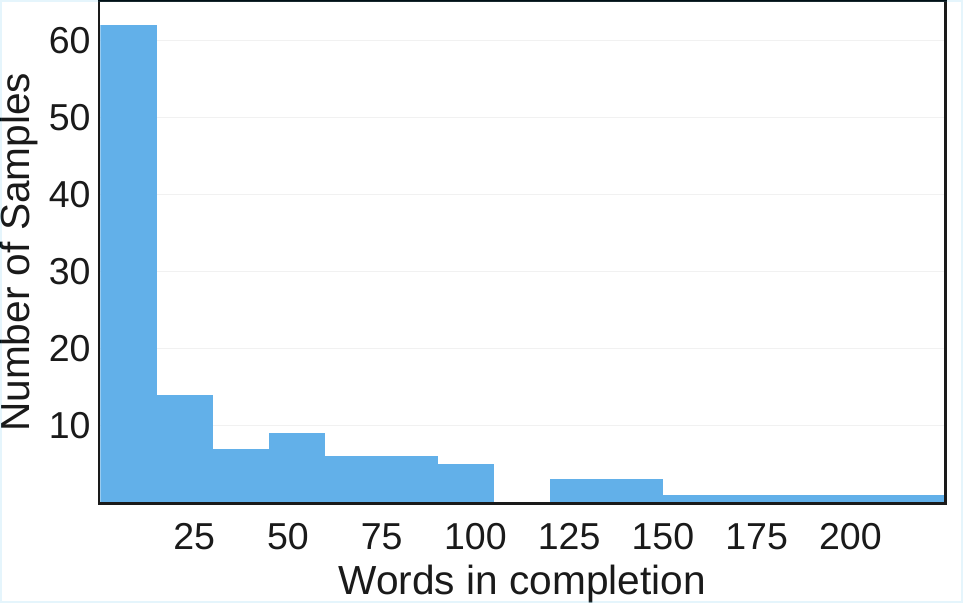}
    \caption{Sample distribution in \textsc{FormaMentis} by number of words in a completion. We measure the completion length as the average number of words between both completions in a sample.}
    \label{fig:completions}
\end{figure}

\end{document}